# Revising the Structure of Recurrent Neural Networks to Eliminate Numerical Derivatives in Forming Physics-Informed Loss Terms with Respect to Time

Mahyar Jahani-nasab[1], Mohamad Ali Bijarchi [1*]

Department of Mechanical Engineering, Sharif University of Technology, Tehran, Iran[1].

*Corresponding author: bijarchi@sharif.edu

**Abstract:**

Solving unsteady partial differential equations (PDEs) using recurrent neural networks (RNNs) typically requires numerical derivatives between each block of the RNN to form the physics informed loss function. However, this introduces the complexities of numerical derivatives into the training process of these models. In this study, we propose modifying the structure of the traditional RNN to enable the prediction of each block over a time interval, making it possible to calculate the derivative of the output with respect to time using the backpropagation algorithm. To achieve this, the time intervals of these blocks are overlapped, defining a mutual loss function between them. Additionally, the employment of conditional hidden states enables us to achieve a unique solution for each block. The forget factor is utilized to control the influence of the conditional hidden state on the prediction of the subsequent block. This new model, termed the Mutual Interval RNN (MI-RNN), is applied to solve three different benchmarks: the Burgers' equation, unsteady heat conduction in an irregular domain, and the Green's vortex problem. Our results demonstrate that MI-RNN can find the exact solution more accurately compared to existing RNN models. For instance, in the second problem, MI-RNN achieved one order of magnitude less relative error compared to the RNN model with numerical derivatives.

## 1. Introduction

The Physics-Informed Neural Networks (PINNs) paradigm, initially proposed by Maziar Raissi et al. [1], has been extensively researched to refine and expand the model for more complex problems [2-5]. This research encompasses various training strategies [6-9] and the implementation of diverse deep learning models tailored to specific challenges [10-12]. The efficacy of these approaches often depends on the problem context. For instance, Solera-Rico et al. integrated β-VAE and transformer-based models to encode fluid-flow velocity fields and learn a Reduced Order Model of its spatio-temporal dynamics. This method proved robust in capturing complex temporal dependencies and long-term trends in both periodic and chaotic flow cases, surpassing other Machine Learning (ML) temporal models in prediction accuracy [13]. These advancements have spurred further research into developing new models for a range of partial differential equations (PDEs) [14-17].

Recurrent Neural Networks (RNNs) are designed to identify patterns in sequential data [18]. Unlike feed-forward neural networks, RNNs possess a hidden state or memory to process time-dependent data [19]. This capability makes RNNs suitable for addressing unsteady PDEs, which are inherently sequential as their solutions evolve over time [20]. Viana et al. utilized RNNs for numerical integration of ordinary differential equations, representing physics-informed kernels as directed graphs to estimate missing physics in the original model [21]. While RNNs are efficient with available data [21], solving unsteady PDEs in a physics-informed manner necessitates numerical derivatives. Hu et al. introduced Neural-PDE, a deep learning framework inspired by numerical PDE schemes and Long Short-Term Memory (LSTM) networks, capable of simulating multi-dimensional governing laws represented by time-dependent PDEs and predicting data for subsequent time steps with high accuracy [22]. Similarly, Liang et al. proposed a Physics-Informed Recurrent Neural Network (PIRNN) to solve time-dependent PDEs, comprising multiple PIRNN cells with input hidden layers, LSTM cells, and output hidden layers. Their approach achieved an accuracy an order of magnitude higher than PINNs [23].

However, employing numerical derivatives introduces several challenges during the training process. This method provides the function's value only at specific points [24], necessitating additional methods like interpolation to estimate values between these points [25]. This limitation underscores the inherent difficulties in using numerical derivatives within the context of RNNs and unsteady problem-solving, as it hinders the generalization capabilities of Artificial Intelligence (AI) [26]. Furthermore, calculating numerical derivatives in RNN structures to form loss functions requires increasing the number of blocks, which can lead to issues such as vanishing gradients [27-29]. Additionally, selecting an appropriate finite difference approach, whether first-order or second-order, adds another layer of complexity [30-31]. The inherent error associated with these methods, arising from the approximation used in the finite difference method, can result in inaccuracies in the calculated derivative values. Moreover, the choice of the finite difference scheme—forward, backward, or central—can also affect the accuracy of the derivative approximation [32-34].

The challenges associated with employing numerical derivatives for training RNNs motivated us to propose a novel approach, termed Mutual Intervals Recurrent Neural Networks (MI-RNN), to effectively solve unsteady PDEs without the need for numerical derivative calculations. Our methodology introduces several key modifications to enable the use of time as an input for direct derivative calculation during the backward process. In Section 2.1, we introduce the concept of mutual information to maintain continuity in derivatives. Our contribution also includes the introduction of a conditional hidden state, detailed in Section 2.2. Furthermore, Section 2.3 outlines the development of a novel loss function and forget factor to facilitate the training process. The impact of each modification is thoroughly investigated in Section 3.1, with MI-RNN studied in terms of accuracy (Section 3.2), and robustness (Section 3.3) for different PDEs.

1. Methodology

Figure 1 provides an overview of traditional Recurrent Neural Networks (RNNs) (a) and our proposed structure (b). Traditional RNNs rely on numerical derivatives between blocks to solve unsteady partial differential equations (PDEs) [35]. However, to make the loss function completely physics-informed without using numerical derivatives, several challenges can arise, which we discuss in the following sections.

In this research, we present an innovative methodology that eliminates the need for numerical derivatives in RNNs. Our approach, termed the 'Mutual Intervals RNN', modifies the RNN structure to incorporate time as an input for each block. The primary advantage of this method is its ability to solve unsteady PDEs without requiring numerical derivatives to form the loss function. To address the challenges associated with using time as an input, we introduce several modifications. Firstly, the mutual loss block is defined over the mutual time intervals to address the lack of sufficient boundary conditions. Secondly, the hidden state is conditioned on time to ensure the uniqueness of the predictions. Lastly, forget factors (FF) are employed to enhance the effectiveness of the hidden states. These modifications, discussed in detail in the following sections, enable the practical use of time as an input for the RNN.

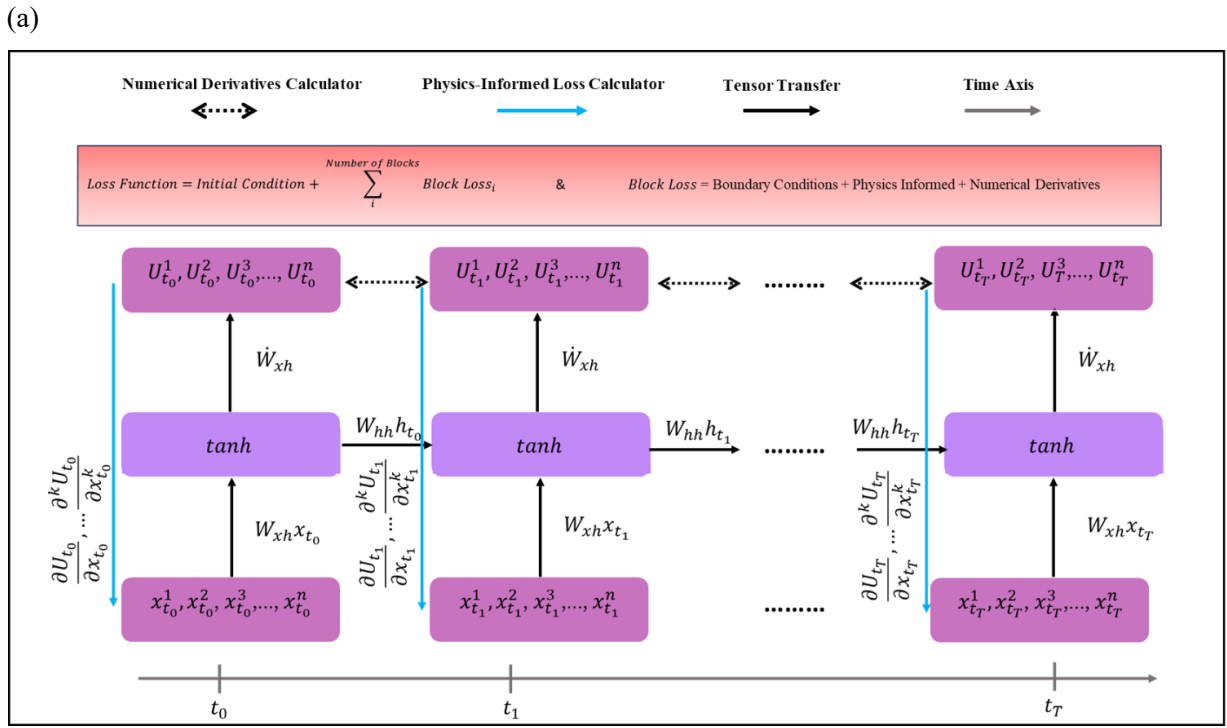

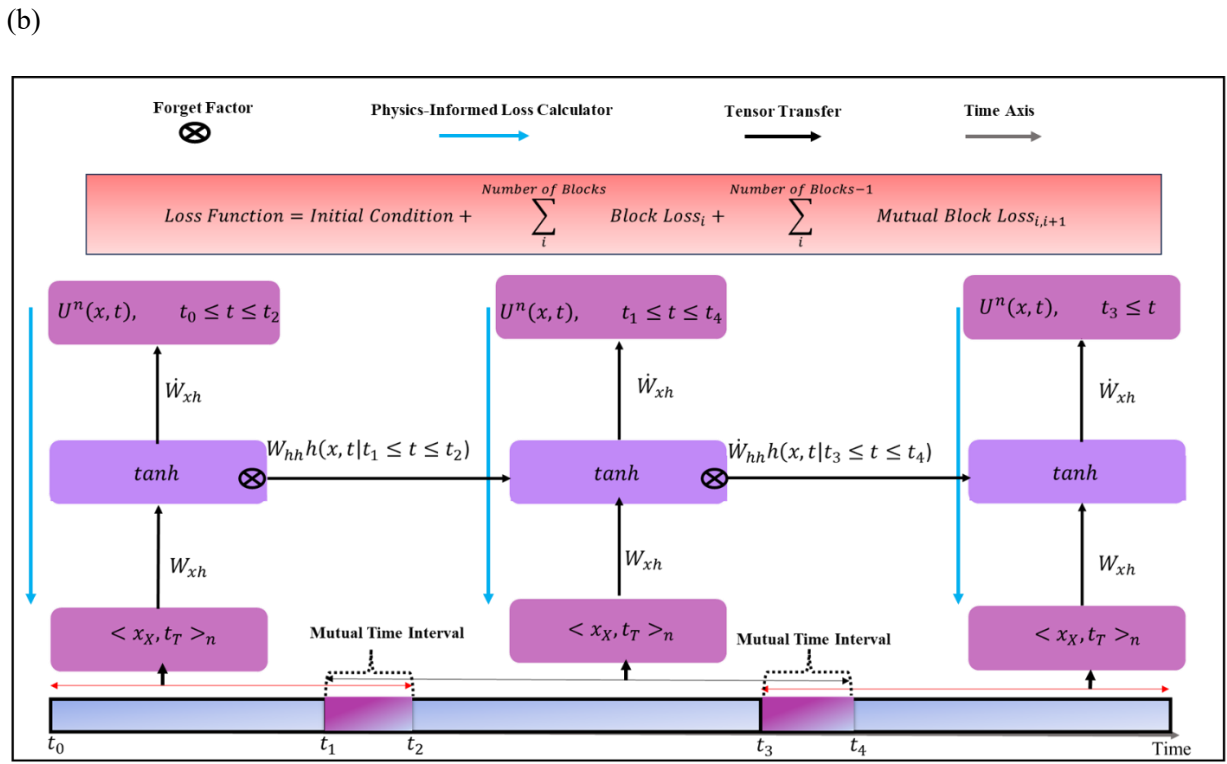

Figure 1 Traditional RNNs for solving PDEs using numerical derivatives with shared weight (a) MI-RNN structure with three blocks and shared weights (b)

## 2.1 Mutual Intervals

As illustrated in Figure 1, the first modification we implemented in the RNN structure involves each block predicting a mutual time interval with its preceding block. This overlap, referred to as the 'mutual time intervals', means that the same state can be predicted in two adjacent RNN blocks. This redundancy enhances the continuity and robustness of the predictions made by each block. It's important to note that the number of the mutual intervals equals the number of the RNN blocks minus one.

Following the removal of numerical derivatives, each block of the RNN is dedicated to solving the PDE within the designated interval. Consequently, each block necessitates its distinct initial condition, which is currently only accessible for the initial block. A novel modification in the structure of the MI-RNN has been introduced, which involves the incorporation of mutual time intervals. This innovative change facilitates the formation of separate initial conditions for each block. This is achieved by defining a new loss term, which is based on the Mean Squared Error (MSE) criteria. The definition of this new loss term is as follows: $\frac{1}{N}\sum[U^i_{(x,t_1 \leq t \leq t_2)}{}^2 - U^{i+1}_{(x,t_1 \leq t \leq t_2)}{}^2]$ where i represents the number of each block. Note that this loss term uses obtained predictions from the previous block as the initial conditions for the next block in the mutual time interval. To observe this modification in action, consider the Burger's equation as follows: $f_0 = +\frac{\partial u}{\partial t} + u\frac{\partial u}{\partial x} - \mu\frac{\partial^2 u}{\partial x^2}$. The exact solution is given by: $u(x,t) = \frac{2 \times 0.01 \times \pi \times \sin(\pi x) \times e^{-0.01\pi^2 \times (t-5)}}{2 + \cos(\pi x) \times e^{-0.01\pi^2 \times (t-5)}}$. The parameters are defined within the range of $0 \leq x \leq 4, and\ 0 \leq t \leq 5$ with $\mu = 0.01\ \frac{kg}{m.sec}$.

In order to solve this problem using MI-RNN, the time interval is partitioned into three segments: the initial segment spanning from 0 to 2.5 seconds, the mutual segment from 2.5 to 2.6 seconds, and the final block which forecasts the output for the time interval of 2.6 to 5 seconds. For the mutual segment, both blocks are capable of predicting the output due to the defined MSE loss function. This is demonstrated in Figure 2, where both blocks are able to accurately predict the solution for time = 2.53s. It is important to note that these predictions within the mutual interval serve as the initial conditions for the solution of the PDE in the second block.

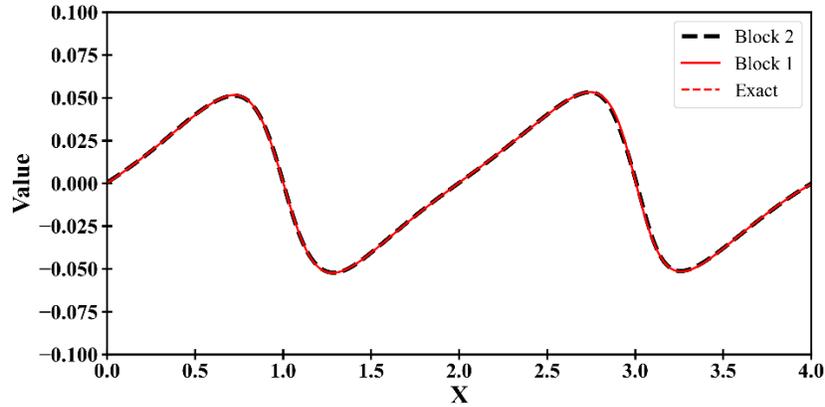

Figure 2 Predictions of MI-RNN in a mutual interval.

## 2.2 Conditional Hidden States

Using time as input in a conventional RNN with standard hidden states can lead to non-unique predictions. The reason is that the hidden state at any given time step is influenced not only by the current time input but also by the sequence of past time inputs. As a result, even if the current time input is identical, the hidden state (and thus the prediction) can vary depending on the sequence of past time inputs. To comprehend this, it's beneficial to examine the initial block of the RNN, which operates independently without any influence from hidden states. This design is intentional and allows it to predict the solution of the PDE for the defined time interval, similar to a Multi-Layer Perceptron (MLP). However, for subsequent blocks, the predictions are influenced by the hidden states. For instance, consider an RNN with two blocks that use time as an input. The second block, for a fixed set of inputs (x, t), can predict different results depending on the information transferred by the hidden state. The hidden state is also a function of the input of the first block of the RNN. This means that if the time is changed for the input of the first block, the predictions of the second block are also influenced. This introduces a level of complexity: due to the problem introduced, all blocks in the RNN, except for the first one, are unable to predict a unique solution. Instead, they generate varying outcomes based on the input from the preceding block. This implies that the solution provided by a block is dependent on the input time used in its preceding blocks. As a result of this, the sequential nature of predictions is questioned, and the nature of the RNN is compromised.

This issue is clearly depicted in Figure 2, which demonstrates that altering the time input for the first block results in varying predictions for the second block at $t = 3.5\ s$. In this figure, the yellow region represents the range of possible variations in the second block's predictions when the time input for the first block is adjusted between 0 and 2.5 seconds. The dotted grey line represents the average value of all the second block's predictions for a given

$x$ and $t = 3.5\ s$, while the red line illustrates the exact solution of Burger's equation at this specific moment. In other words, if the second block is tasked with predicting the output for a pair ($x, t = 3.5\ s$), it can produce different solutions based on the time value used in the first block. This problem intensifies for the subsequent blocks, which are influenced by all their preceding blocks, leading to a significant diversity in predictions. This highlights the intricate interdependencies within recurrent neural networks (RNNs) when time is used as an input. However, this is a unique use case for solving unsteady partial differential equations (PDEs),, and the inherent characteristics of RNNs are being adapted to achieve distinct solutions.

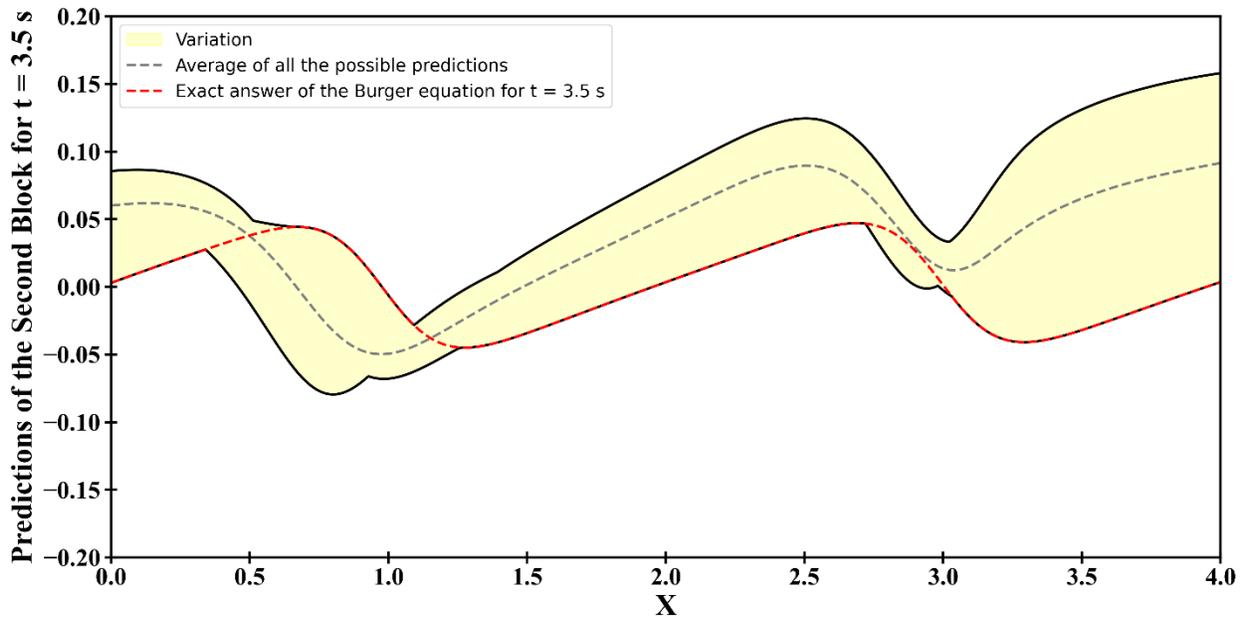

Figure 3: Predictions of the second block of the RNN when different times are used as inputs for the first block.

We addressed the problem by making the prediction of each block in the MI-RNN dependent on a particular moment from the preceding block both in the training and inference phases. Consider the first two blocks of an MI-RNN as shown in Figure 2. The first block predicts the output for the time interval $t_0 \leq t \leq t_2$, while the second block predicts the output for the interval $t_1 \leq t \leq t_3$. The mutual time interval $t_1 \leq t \leq t_2$ can be predicted by both blocks as discussed in the previous section. For clarity, the predictions are categorized into three parts: the mutual interval which can be predicted by both blocks, and the independent intervals predicted by each block ($t_0 \leq t < t_1$ and $t_2 < t \leq t_3$). Therefore, for predicting outputs in the time interval $t_2 < t \leq t_3$ by the second block, we condition the hidden state on the most recent information from the first block at $t_2$. To make the hidden state conditional, Block 1 takes an input at time $t_2$, and produces an output. This output is then passed as the hidden state to Block 2. Now, when Block 2 is making its prediction at time $t_2 + t$, it considers not only its own input at time $t_2 + t$ but also the hidden state passed from Block 1 at $t_2$. This hidden state encapsulates information from the input at time $t_2$ in Block 1. By using a fixed input for Block 1, we ensure that the hidden state passed to Block 2 is conditioned on this specific input. This means that the prediction made by Block 2 at time $t_2 + t$ is dependent on the input at time $t_2$, in Block 1. This approach aligns the predictions made by different blocks because the prediction at any given time step is only determined by the current input and a specific moment from the preceding block. This leads to unique and consistent predictions across different blocks in the MI-RNN. Additionally, it's crucial to understand that the choice of time for the condition is guided by the fact that the most recent prediction from the first block offers the most relevant information for the second block's predictions. Now for the mutual time interval $t_1 \leq t \leq t_2$, both blocks have the ability to predict the output. In this situation, if the second block is predicting for a time $t_m$ within the mutual interval, the first block should also use $t_m$ as its time input. By conditioning the hidden state on the current time, the hidden state, which contains information from previous inputs, helps align the predictions made by different blocks of the MI-RNN. This ensures consistency in predictions when two blocks are predicting the output for the same time point.

**2.3 Loss Function and Forget Factor**

The final modification involves the use of a coefficient to regulate the influence of the hidden states. This coefficient, referred to as the 'forget factor' (FF) in this study, is a number between 0 and 1 that determines the significance of the hidden state on the predictions of each block. The use of an FF is vital for our model due to the following issue. In certain instances, the information transferred is not always advantageous. Consider a scenario where the first block of the RNN predicts the solution for the time interval $t_0 \leq t \leq t_1$, while the second block predicts the solution for $t_1 < t \leq t_2$. If the hidden state transfers data related to $t_1$ to the second block, only the predictions for $t_1 < t \leq t_1 + \varepsilon$ will benefit from this hidden state, assuming that $\varepsilon$ is sufficiently small. This is due to the hypothesis that in unsteady problems, the solutions at times that are close together in the temporal dimension often exhibit a high degree of similarity. However, it's important to note that this is a general tendency, and if it's not true the value of the FF can be simply changed. Lastly, we define the loss function for the MI-RNN, which will be used for training our proposed model. As shown in Figure 2, the loss function is defined as follows:

$$Loss\ Funciton = Initial\ Condition + \sum_{i}^{Number\ of\ Blocks} Block\ Loss_i + \sum_{i}^{Number\ of\ Blocks-1} Mutual\ Block\ Loss_{i,i+1}$$

The loss function of MI-RNNs for unsteady problems is composed of three distinct parts. The first term represents the initial condition, an essential component in unsteady problems. This term is calculated by the first block of the MI-RNN, which provides domain points at the starting time and compares the predictions with the true values for this time step, forming this loss term.

The second term, known as the Block Loss, is used to compute two distinct components for each block: boundary conditions and PDE loss (physics-informed loss [1]). The PDE loss guarantees compliance with the fundamental physical laws associated with the problem at hand. It's important to note that the PDE loss is derived through the computation of output derivatives from the given inputs. For instance, in the context of Burger's equation discussed in Section 3.1, each block is characterized by two inputs ($x$ and $t$) and one output ($u$). Therefore, the physics-informed loss, represented as $f_0$, is directly computed.

The final term of the loss function, referred to as the Mutual Block Loss, consists of one primary component and two optional components. The primary component, which is essential for meaningful predictions, aims to establish a connection between the blocks. As detailed in Section 3.2, this loss term is formulated by employing the MSE loss between the prediction of one block and its subsequent block over the mutual interval. The remaining two components of the loss function pertain to the boundary conditions and the physics-informed loss. These components, while not as crucial as the primary component, can enhance the accuracy of the predictions when formulated over the mutual time interval between each pair of MI-RNN blocks, taking into consideration the conditional hidden state as discussed in the preceding section.

Finally, with time serving as the input, the loss function becomes purely physics-informed, thereby eliminating reliance on numerical derivatives. This approach ensures that the model is grounded in the underlying physical principles of the problem, enhancing its robustness and reliability.

## 3 Results and Discussion

### 3.1 Parameter analysis

In this section, we delve into the modifications applied to the structure of the MI-RNN, utilizing the Burger equation as outlined in Section 3.2. As per this equation, each block within the MI-RNN is designed to have two inputs and one output. Each of these blocks is composed of four layers, with each layer housing 30 neurons. It is crucial to highlight that the weights are shared across all blocks of the MI-RNN. Each block contains four hidden states, and each of these hidden states comprises 30 neurons. The activation function used consistently across all benchmarks is the hyperbolic tangent (tanh), and the Adam algorithm is utilized as the optimizer. The accuracy of the model is measured using the R-squared metric, calculated using the formula $1 - \frac{Residual\ Sum\ of\ Squares}{Total\ Sum\ of\ Squares}$. To evaluate the accuracy during the inference phase, we generate a set of points with a spacing of 0.01 in both the x and t dimensions. These points are subsequently used to create a mesh grid. This grid, consisting of 200,000 input points, serves as the input for our model. This allows us to effectively assess its performance and accuracy by comparing the model's predictions against the exact solution. It is important to note that the features mentioned in this subsection are consistent throughout.

Firstly, we examine the influence of two critical factors on the accuracy of the MI-RNN: the duration of the time intervals and the number of MI-RNN blocks. To establish mutual time intervals, we initially divide the time range of $0 \leq t \leq 5$ into various intervals, the count of which corresponds to the number of blocks (as depicted in Figure 4). This figure illustrates the impact of the mutual block's duration on independent predictions (In this context, the independent part refers to the segment of the time interval where only one block is capable of making predictions). When the number of blocks in the RNN is small, employing a longer mutual time interval undermines the role of the independent part of the time interval. It's important to note that for both Table 1 and Figure 4, the hidden state is conditioned on the last moment of the previous block's mutual time interval, with an FF value of 1 for both independent and mutual time intervals. The final row of Figure 4 highlights a significant issue: using a large mutual time interval compromises the effectiveness of an MI-RNN block. In this row, the mutual time interval (purple) completely overlaps with the independent time interval (blue) of the adjacent blocks, rendering some of the blocks trivial.

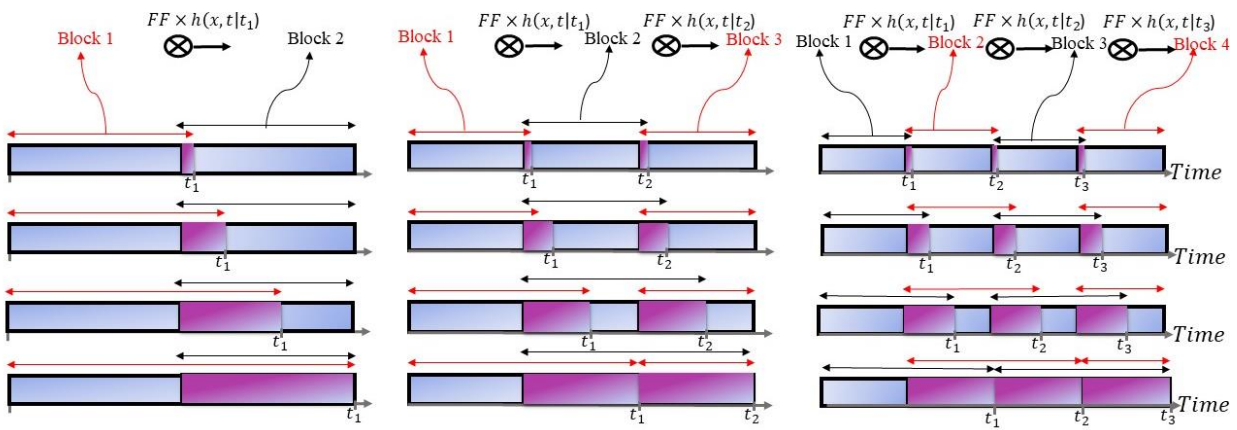

Figure 4 The mutual intervals are represented by the color purple, while the independent parts of the interval are shown in blue. The hidden state (h) of the first block is conditioned on $t_1$, and the subsequent hidden states are conditioned on $t_2$ and $t_3$, respectively.

The role of the length of the mutual interval in the accuracy of the MI-RNN for solving Burger's equation is explored in Table 1. The duration of the time domain, set at 5 seconds, is divided into a total number of blocks to establish time intervals. For example, in the last row of Table 1, the initial division of intervals is as follows: $0 \leq$ First Block $\leq \frac{5}{4}, \frac{5}{4} \leq$ Second Block $\leq 2 \times \frac{5}{4}, 2 \times \frac{5}{4} \leq$ Third Block $\leq 3 \times \frac{5}{4}$, and $3 \times \frac{5}{4} \leq$ Fourth Block $\leq 4 \times \frac{5}{4}$. After this initial division, we adjust the time duration for each block based on the mutual time intervals, as detailed in Table 1. Consider the last row of the last column where the mutual time interval is 1 second. In this scenario, the time intervals for each block are adjusted to: $0 \leq$ First Block $\leq \frac{5}{4} + 1, \frac{5}{4} \leq$ Second Block $\leq 2 \times \frac{5}{4} + 1, 2 \times \frac{5}{4} \leq$ Third Block $\leq 3 \times \frac{5}{4} + 1$, and $3 \times \frac{5}{4} \leq$ Fourth Block $\leq 4 \times \frac{5}{4}$. Note that the end of each block is delayed by 1 second. This configuration results in overlapping time intervals between successive blocks. However, this overlap is intentional and critical for the model's performance. It facilitates connections between blocks, with both blocks predicting the same output based on the defined loss function. It is important to note that the length of each interval can be adjusted based on specific design requirements, and they do not necessarily have to be of equal length.

The R-squared metric, as presented in Table 1, provides a measure of the accuracy of each model. A closer examination of the first column reveals a significant decrease in accuracy when the mutual time intervals are not incorporated into the model. For instance, a model with two blocks and without a mutual interval yields an accuracy of 0.8012. However, the accuracy decreases substantially as the number of blocks increases. n this particular case, the absence of mutual time intervals renders the other blocks ineffective. This is because no initial condition is provided for this block, resulting in an insufficient number of boundary conditions to accurately predict the solution of the PDE for their time interval. This underscores the crucial role of mutual time intervals in the MI-RNN structure.

While the mutual time interval is essential for this model, extending its duration does not enhance prediction accuracy, as shown in Table 1. If the interval is too long, adjacent blocks predict the same outcome, making one redundant. This inefficiency wastes resources without additional benefits. Therefore, based on Table 1, the mutual time interval should be kept short. For instance, a mutual time interval of 0.01 is sufficient, and adjusting other hyperparameters may be more effective for improving model accuracy.

Table 1: Accuracy of the MI-RNN for different numbers of blocks and various lengths of mutual time intervals

|   | Not Employed | 0.01 | 0.5 | 1 |
|---|---|---|---|---|
| 2 | 0.8012 | 0.9979 | 0.9997 | 0.9988 |
| 3 | 0.6054 | 0.9995 | 0.9150 | 0.9726 |
| 4 | 0.5815 | 0.9980 | 0.9067 | 0.8967 |

Up to this point, we have used an FF value of 1 in the mutual time interval and 0 in the independent blocks. In this section, we explore different FF values and various hidden states conditioned on different moments.

To delve deeper into the optimal state for each of these factors, we trained an MI-RNN with two blocks and a mutual time interval of 0.05 s to solve Burger's equation. This process is illustrated in Figure 5 which provides a comprehensive representation of the model. The first block is capable of making predictions for the time interval $0 \leq t \leq 2.55$, while the second block makes predictions for the time interval $2.5 \leq t \leq 5$. The independent part of the second interval is further divided into two sub-intervals: $2.55 \leq t \leq 2.6$ and $2.6 \leq t \leq 5$. This division is specifically implemented in this case to examine the influence of the FF within the mutual time interval, an independent time interval near the mutual time interval, and the rest of the independent time interval. As depicted in Figure 5, we can condition the hidden states at different moments for each time interval. This approach allows us to observe the impact of the FF and the conditioned hidden state on the model's predictions.

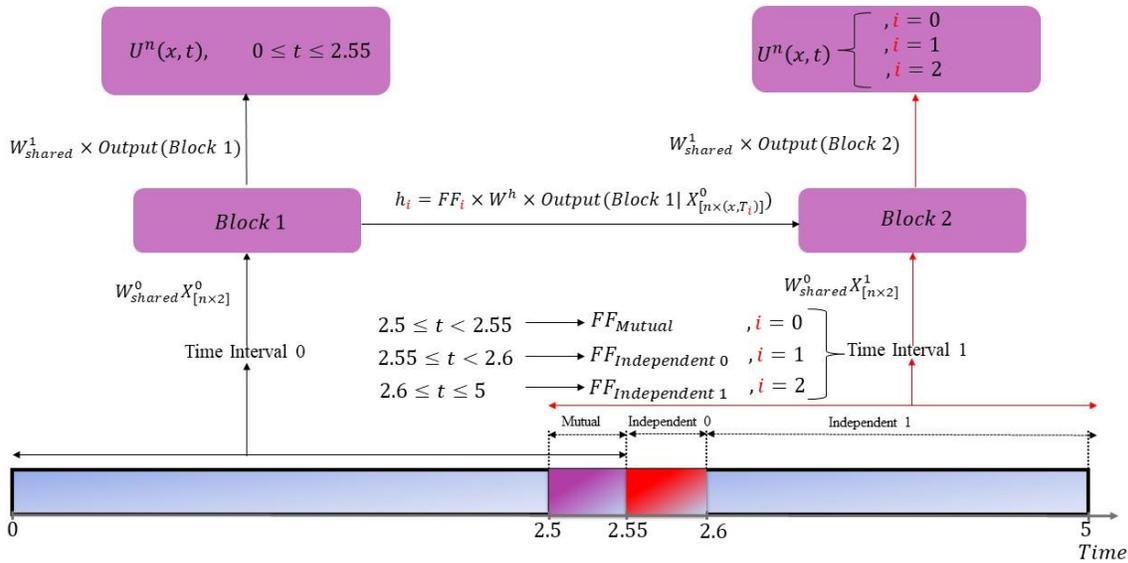

Figure 5 Illustration of the MI-RNN incorporating the Forget Factor (FF) and Conditional Hidden State (CHS) in different time interval of the second block.

Table 2 shows the R-squared accuracy of the model depicted in Figure 5, where the hidden state is conditioned at different moments with varying FF values. The time interval of the second block is divided into three parts: the mutual interval and two independent intervals. For each of these intervals, the hidden state is conditioned at different moments, represented by the rows, with an FF value shown in the columns. In this table, TA refers to temporal alignment, meaning the hidden state of the second block is conditioned on the same time step for which the second block is making a prediction. In other words, if the second block's output is for (x, t = n), the same input is provided to the first block. Therefore, when the predictions of the second block extend beyond the mutual interval, the hidden state is conditioned on a moment outside the training interval of the first block. In rows of Table 2, three distinct fixed times are considered, indicating the hidden state is conditioned at these times: within the mutual interval (2.55), at the start of the first interval (0), and outside the first block's interval (2.7). The columns of Table 2 represent the FFs. For instance, FFs = [1,0,0.1], shows that the hidden state's information for the mutual time interval is transferred completely (1×hidden state), while for the first independent time interval, no information is transferred (0×hidden state). For the second independent time interval, only 0.1 of the total information is transferred (0.1×hidden state).

The data in Table 2 reveals intriguing insights about the relationship between different conditional hidden states, FFs, and their corresponding accuracies. The third row, denoted by [2.55, 2.55, 2.55], exhibits the highest overall average accuracy of 0.9674, calculated by averaging all the accuracies in that particular row. This suggests that conditioning the hidden state at the last moment of the training interval yields the most accurate results on average. However, not all moments are equally beneficial for conditioning. For instance, the scenarios [TA, 2.7, 2.7] and [TA, 0, 0] result in relatively lower outcomes. This could be attributed to the fact that the hidden state is conditioned on a moment outside its training interval or a moment that is far from the interval of the second block. When the FF is set to zero, the model's predictive accuracy experiences an improvement, only in cases where the chosen moment for conditioning is not optimal. For example, when the hidden state of the last independent interval is conditioned on a moment that does not contribute positively to the second block, setting the FF to zero effectively neutralizes its impact, thereby enhancing the accuracy. This pattern is evident in the second row and the final four rows of Table 2. This observation underscores the importance of judiciously selecting the moments for conditioning and the value of FF, as they can influence the model's performance. In conclusion, the third row of the table, with the accuracy of 0.9999 in that row and FF values of [0.5,0.5,0.5], demonstrates the best performance. This suggests that the optimal strategy is to condition the hidden state on the final moment of the training interval, initially setting the FFs to 0.5 within the interval. Subsequently, the FF is adjusted for that specific interval across different PDEs to evaluate the utility of the conditioned hidden state.

Table 2: R-squared values for various conditional hidden states and FFs are reported. Note that a negative R-squared, as per the formula, indicates a model performing worse than a simple horizontal line.

|  | [1,0.5,1] | [1,0.5,0.5] | [0.5,0.5,0.5] | [1,0.1,0] | [1,0,0.1] | [1,1,1] |
|---|---|---|---|---|---|---|
| [ TA, TA , TA] | 0.9632 | 0.5934 | 0.9886 | 0.9845 | 0.7676 | 0.9739 |
| [ TA, TA , 2.55] | 0.3546 | 0.5974 | 0.7737 | 0.9844 | 0.7898 | 0.3118 |
| [2.55, 2.55 , 2.55] | 0.9728 | 0.8477 | 0.9999 | 0.9992 | 0.9914 | 0.9935 |
| [ TA, 2.55 , TA] | 0.9801 | 0.8305 | 0.9749 | 0.9990 | 0.5509 | 0.9510 |
| [ 2.55, 2.55 , TA] | 0.7527 | -0.1868 | 0.8132 | 0.9995 | 0.9039 | 0.5898 |
| [ TA, 2.7 , 2.7] | 0.2157 | 0.2565 | 0.6563 | 0.9987 | 0.6811 | 0.2067 |
| [ TA, 0 , 0] | 0.6915 | 0.3025 | -0.5432 | 0.9870 | 0.8255 | 0.6933 |

### 3.2 Accuracy

In this section, we delve into the exploration of the MI-RNN model's proficiency in solving a variety of PDEs. Initially, we scrutinize the accuracy of the MI-RNN model in solving the Burger's equation, a fundamental nonlinear PDE. This examination provides a deeper understanding of the model's capability in handling complex mathematical problems. Subsequently, the MI-RNN model is juxtaposed with traditional RNN structures, particularly focusing on those that employ numerical derivatives in the formation of the loss function (unsteady heat conduction). This comparison aims to highlight the distinctive advantages of the MI-RNN model over conventional methods. Lastly, the MI-RNN model is benchmarked against more innovative approaches for solving the unsteady Navier-Stokes equations, specifically the Taylor Green Vortex problem.

#### 3.2.1 Burger

In some studies, it has been shown that the MSE increases over time when using MLP-based models, which is a drawback that attracts people to RNNs [32]. As depicted in Figure 6, the MSE of the MI-RNN with four blocks is analyzed, with a mutual interval of 0.01 for the Burger equation. Interestingly, the MSE remains relatively consistent during the transition from one block to another. The observation of a low and consistently maintained error over time suggests that the MI-RNN is a highly accurate model for unsteady problems. In all figures, blue represents the independent interval, while purple indicates the mutual interval.

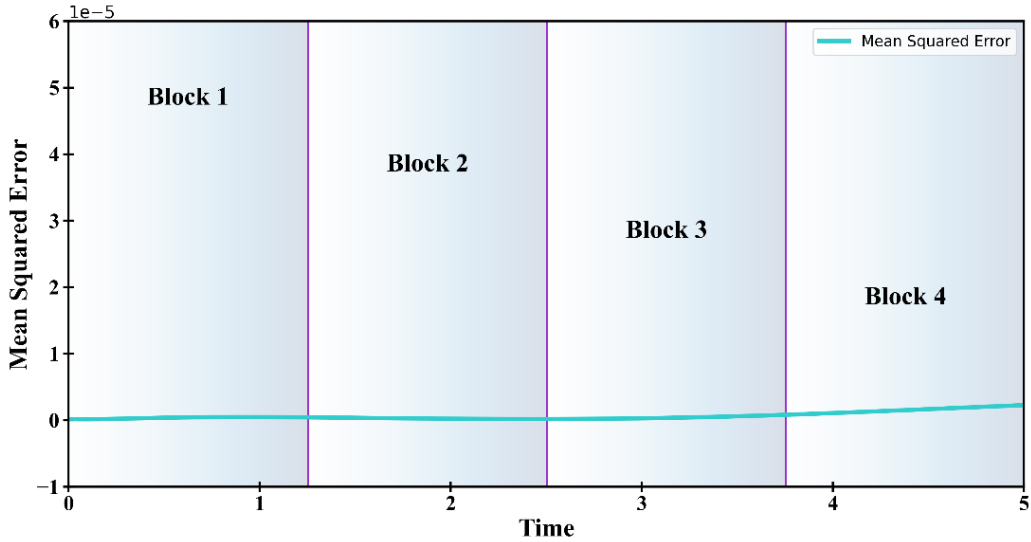

Figure 6 The Mean Squared Error (MSE) over time for the Burger's equation.

The accuracy of the MI-RNN model is further examined within the domain. Figure 7(a) illustrates the predicted values by MI-RNN, the exact values of the equation, and the absolute error between these two figures. Figure 7(b) compares the exact solution of the equation with the first block of the RNN, then within a mutual interval, and finally within the second interval for an MI-RNN with two blocks. For this model, the hidden state is conditioned on 2.55, and the gate factor is set to 0.5, achieving an R-squared accuracy of 0.9999 as reported in Table 2. Furthermore, Figure 7(c) demonstrates that when the solution lies within the first independent interval, the first block (represented in teal) accurately predicts it. Conversely, when the solution is in the second independent interval, block 2 can predict it. It is important to note that within each independent block, only one block is capable of predicting the solution. However, near the mutual interval, as illustrated in purple, both blocks predict the solution accurately.

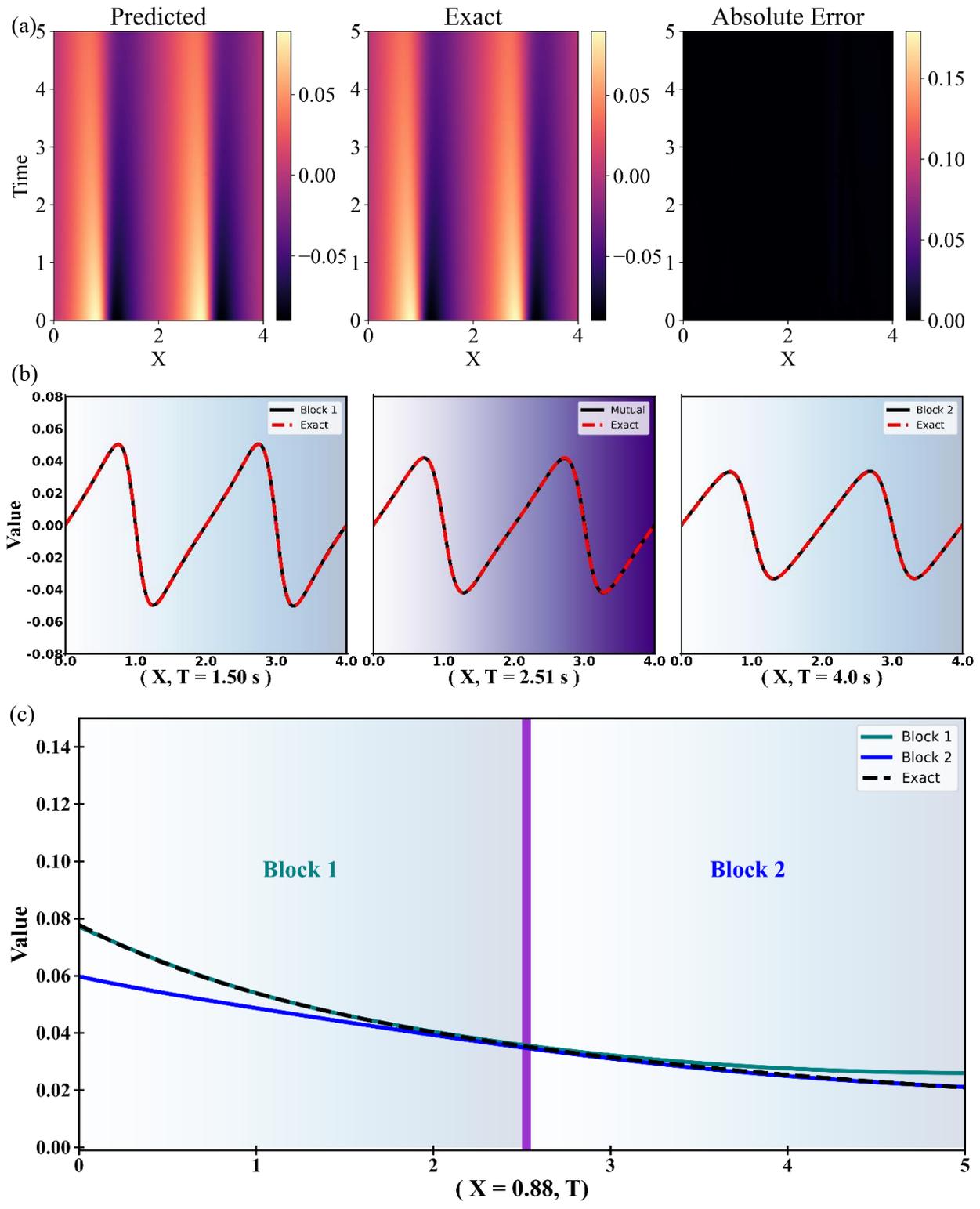

Figure 7: (a) An illustrative comparison between the predicted and exact solutions of Burger equation. (b) A juxtaposition of the first block's prediction with the exact solution for time steps 1.50, 2.51, and 4.0, displayed from left to right. (c) A similar comparison over time with a fixed X value of 0.88.

## 3.2.2 Unsteady Heat Conduction

In Section 2, we discussed the necessity of calculating output derivatives with respect to time using numerical derivatives when the MI-RNN is not employed. Liang et al. incorporated an LSTM cell in each block of their model, and by using numerical derivatives for time, they formed the loss function and solved the following equation for an irregular domain: $\frac{\partial u}{\partial t} = -\frac{1}{2}(\frac{\partial^2 u}{\partial x^2} + \frac{\partial^2 u}{\partial y^2})$ [35]. They then compared their proposed model with MLP-based PINN ($x$, $t$ as inputs) in an irregular domain and achieved an error order less than that of MLP-based PINN. We trained the MI-RNN under the same domain and conditions as those in reference [35] (for more details, refer to the first benchmark of reference [35]). We compared the accuracy of our model with the same metric used in this study [35] on the time steps they reported. The modifications implemented in the MI-RNN, eliminate the numerical derivatives from the RNN's loss function, provide our model with two advantages over models that use numerical derivatives. First, it can predict the answer with higher accuracy. Secondly, unlike traditional RNNs that use numerical derivatives to calculate time derivatives and require additional methods like interpolation to predict outputs for time steps between their training, the MI-RNN model can predict the output directly. The generalization capability of the MI-RNN model is a feature that sets it apart from models using numerical derivatives. To demonstrate the superior accuracy of the MI-RNN model, we intentionally omitted time steps reported in Table 3 from the training input and used the model's generalization capability to calculate the accuracy. This means that even though these time steps were not used during the training process, the MI-RNN model was still able to maintain higher accuracy. It is important to note that the accuracy of the model is determined by comparing its predictions with the exact solution of the PDE. This exact solution is reported in [35] and is given by the following expression: $u(x, t, y) = e^t \sin x \sin y$.

Table 3 An examination of the relative error between the MI-RNN and the results presented in reference [35]. Note that t = 3×Δt.

| Δt | MI-RNN | PINN [35] | PIRNN[35] |
|---|---|---|---|
| 0.1 | 4.4061×10$^{-4}$ | 2.5220e×10$^{-2}$ | 4.3787×10$^{-3}$ |
| 0.05 | 3.5507×10$^{-4}$ | 5.5564×10$^{-2}$ | 9.4351×10$^{-3}$ |
| 0.01 | 2.4406×10$^{-4}$ | 4.7063×10$^{-2}$ | 5.9227×10$^{-3}$ |
| 0.001 | 2.7664×10$^{-4}$ | 3.6436×10$^{-2}$ | 6.9781×10$^{-3}$ |
| 0.0001 | 2.8176×10$^{-4}$ | 3.9214×10$^{-2}$ | 6.4166×10$^{-3}$ |

### 3.2.3 Taylor Green Vortex

In fluid dynamics, the Taylor-Green vortex represents an unsteady, decaying vortex flow with an exact solution to the incompressible Navier-Stokes equations in Cartesian coordinates. This problem is utilized to examine vortex dynamics, turbulent transitions, and energy dissipation processes. It is selected to evaluate the accuracy of the MI-RNN for more complex scenarios. The Taylor-Green vortex is defined for x× y× t ϵ [0, 2π] × [0, 2π] × [0, T] (For more details, please refer to [36]).

$$\frac{\partial u}{\partial x} + \frac{\partial v}{\partial y} = 0$$

$$\frac{\partial u}{\partial t} + u\frac{\partial u}{\partial x} + v\frac{\partial u}{\partial y} = -\frac{1}{\rho}\frac{\partial p}{\partial x} + v(\frac{\partial^2 u}{\partial x^2} + \frac{\partial^2 u}{\partial y^2})$$

$$\frac{\partial v}{\partial t} + u\frac{\partial v}{\partial x} + v\frac{\partial v}{\partial y} = -\frac{1}{\rho}\frac{\partial p}{\partial y} + v(\frac{\partial^2 v}{\partial x^2} + \frac{\partial^2 v}{\partial y^2})$$

The MSE for the velocity components (U and V) and the pressure field (P) at different viscosity values was analyzed to assess the model's accuracy in solving the Taylor-Green vortex problem. In Table 4 The MSE was averaged over uniformly sampled time steps. For the velocity components, the model performed better at higher viscosity values. However, for the pressure field, the model performed best at medium viscosity.

Table 4 To report the accuracy, the Mean Squared Error (MSE) is calculated and averaged over 10 uniformly sampled time steps.

|   | $v = 0.01$ | $v = 0.1$ | $v = 1.0$ |
|---|---|---|---|
| U | $2.102 \times 10^{-5}$ | $3.077 \times 10^{-5}$ | $1.500 \times 10^{-5}$ |
| V | $2.24 \times 10^{-5}$ | $4.131 \times 10^{-5}$ | $1.257 \times 10^{-5}$ |
| P | $5.74 \times 10^{-5}$ | $1.818 \times 10^{-5}$ | $7.013 \times 10^{-4}$ |

### 3.3 Robustness

When solving PDEs, especially in the context of inverse problems, the proposed model must demonstrate robustness against noise. In this section, we evaluate the robustness of the MI-RNN model by introducing different levels of Gaussian noise. To achieve this, we add Gaussian noise with a mean value of zero and standard deviations of 1, 0.1, and 0.01 to the predicted values of the first block in the mutual interval. As a result, the mutual loss term defined for the second block is accompanied by an error. After training, we investigate the mean squared error for each block. Figure 9 displays the logarithm of these errors, where higher values indicate better performance.

Remarkably, we observe that even when noise is introduced into the mutual loss function between the blocks, the error is not significantly propagated through the entire structure. This consistency in error across the structure highlights the robustness of the MI-RNN model.

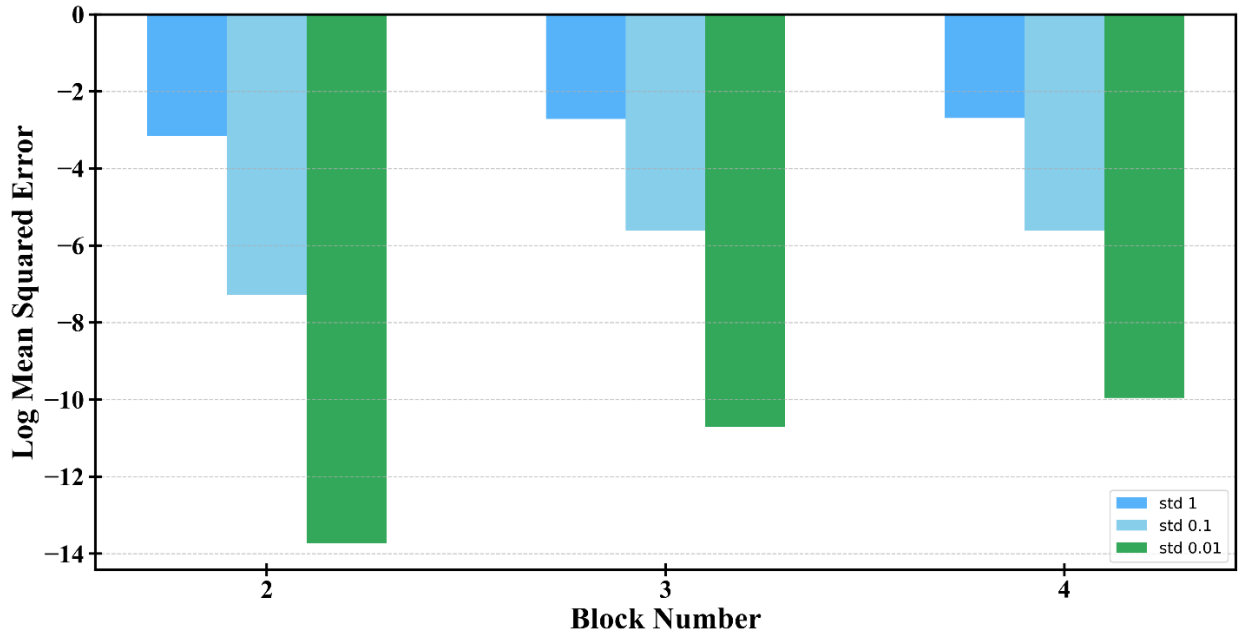

Figure 8 Logarithm mean squared error for each block after adding a noise

## 4. Conclusion

In conclusion, the modifications to the recurrent neural network (RNN) structure have shown promising results in solving unsteady partial differential equations (PDEs) more accurately compared to traditional RNNs and physics-informed neural networks (PINNs). These improvements were demonstrated across three benchmarks: the Burgers' equation, unsteady heat conduction, and the Green vortex problem. By extending the time step of each RNN block and overlapping adjacent blocks, we were able to define a mutual loss function, which eliminated the need for numerical derivatives in forming a physics-informed loss function for traditional RNNs. Furthermore, we employed conditional hidden states to address some problems associated with the extension of the time interval and defined a forget factor for each hidden state. However, the high number of loss terms presents a challenge, potentially leading to an unbalanced training process in our new methodology termed Mutual Interval RNNs (MI-RNNs). Future research will focus on improving this structure and exploring the use of GRU or LSTM layers to address more complex problems, such as two-phase flow.

**Author Contributions:** Methodology & writing, Mahyar Jahani-nasab; review & editing, Mohamad Ali Bijarchi

**Funding:** This research received no external funding.

**Code Availability Statement:** The code and models will be publicly available at [GitHub.](GitHub.)

**Conflicts of Interest:** The authors declare no conflict of interest.